# A Trainable Neuromorphic Integrated Circuit that Exploits Device Mismatch

Chetan Singh Thakur (*Student Member, IEEE*), Runchun Wang, Tara Julia Hamilton (*Member, IEEE*),
Jonathan Tapson (*Member, IEEE*) and André van Schaik (*Fellow, IEEE*)

*Abstract*— **Random device mismatch that arises as a result of scaling of the CMOS (complementary metal-oxide semi-conductor) technology into the deep submicron regime degrades the accuracy of analogue circuits. Methods to combat this increase the complexity of design. We have developed a novel neuromorphic system called a Trainable Analogue Block (TAB), which exploits device mismatch as a means for random projections of the input to a higher dimensional space. The TAB framework is inspired by the principles of neural population coding operating in the biological nervous system. Three neuronal layers, namely input, hidden, and output, constitute the TAB framework, with the number of hidden layer neurons far exceeding the input layer neurons. Here, we present measurement results of the first prototype TAB chip built using a 65nm process technology and show its learning capability for various regression tasks. Our TAB chip exploits inherent randomness and variability arising due to the fabrication process to perform various learning tasks. Additionally, we characterise each neuron and discuss the statistical variability of its tuning curve that arises due to random device mismatch, a desirable property for the learning capability of the TAB. We also discuss the effect of the number of hidden neurons and the resolution of output weights on the accuracy of the learning capability of the TAB.**

*Index Terms*— *Neuromorphic Engineering; Analogue Integrated Circuit Design; Stochastic Electronics; Neural Network Hardware.*

## I. INTRODUCTION

O ver time, electronic devices have witnessed a higher packing density of transistors in accordance with Moore's law [1]. Owing to this, computing devices have become smarter, faster, and more efficient. The increase in the number of transistors has been made possible due to a decrease in the minimum feature size, which has already reduced below 22nm. However, keeping up with Moore's law has not been easy. In submicron technologies, factors such as minor variations of process, external unknown fields, minor layout changes, and leakage currents have large effects on the performance of analogue circuits, making them difficult to design and, thus, creating significant challenges. These effects may be minimised by increasing device size; however, this increases the size of an integrated circuit (IC) and as a result increases its cost [2][3]. Further, the failure of a few transistors may result in the failure of the entire chip, rendering it unusable. Thus, there is a trade-off between performance yield and costs in the submicron technology.

Neuromorphic systems, inspired by neurobiological processing systems, offer an attractive alternative to conventional analogue IC design technology. The brain is an incredible computational device that surpasses today's modern computers in various tasks such as vision and audition. Similar to the problems of transistor failure and device mismatch in an IC, the brain is faced with the problems of heterogeneity of neuronal responses to stimuli and neuronal cell death. The biological nervous system has been able to resolve these problems over the course of evolution, and provides an excellent model for IC implementation.

In many regions of the brain, information is encoded by patterns of activity occurring over populations of neurons, a phenomenon referred to as population coding [4]. We have developed a novel neuromorphic system called a Trainable Analogue Block (TAB) that works in a similar manner by using a large pool of neurons for encoding the input, and linearly combining the neuronal responses to achieve decoding. The TAB chip architecture explicitly uses random device mismatch to its advantage, and is thus ideally suited for submicron technologies. Other research groups have previously suggested the use of random device mismatch in their architectures [5], [6], but, to the best of our knowledge, none of these have yet been tested on ICs. The TAB attempts to incorporate the features of neurobiological systems, such as low power consumption, fault tolerance and adaptive learning. For example, the TAB can use learning to compensate for the failure of a few transistors, thus improving the yield of the system. Owing to adaptive learning, the designs will be portable across technologies and applications, eliminating the need for custom IC design for those functions that can be implemented with our TAB. We envisage that the TAB will contribute to a considerable speed-up in IC design by shortening the design cycle for analogue circuits, and result in a drastic decrease in design costs. The TAB framework may be used to design systems that will employ hardware variability to achieve their engineering goal, thus qualifying as a design circuit paradigm for stochastic electronics [7]. The TAB circuits are effectively universal function approximators [8], thereby allowing for complex processing on a simple and repeatable substrate.

This paper is organised as follows: Section II explains the framework of the TAB. VLSI (Very Large Scale Integration) implementation of the TAB is described in section III, the algorithm setup for offline learning in section IV and constrained algorithm in section V. We present the measurements of the building blocks of our TAB implementation in section VI and conclusions in section VII.

## II. FRAMEWORK OF TAB

The TAB framework draws inspiration from the phenomenon of neural population coding. In population coding, biological neurons in several parts of the brain encode information in a collective and distributed manner using spike rates. The accuracy of information processing in the cortex depends on the quality of population coding, which in turn is affected by the heterogeneity of neuronal responses and the shape of neuronal tuning curves [10][11][10]. The tuning curve of a neuron is a plot of its average firing rate as a function of the input stimulus. As examples of population coding, neurons in monkeys, cricket, barn owl, cats, bats and rats encode the direction of arm movements [12], the direction of a wind stimulus [13], the direction of a sound stimulus [14], saccade direction [15], echo delay [16] and the position of the rat in its environment [17], respectively. The TAB framework is designed to use neuronal tuning curves instead of individual spikes. Further, we have used a heterogeneous population of neurons in the TAB architecture. Heterogeneity of the tuning curves of the neurons increase the encoding capacity of the network [18].

In order to decode the response of a neuronal population, it is required to combine the firing rates of neurons into a population ensemble estimate. Generally, the tuning curve of each neuron contributes a basis function and the best estimate of the physical variables is computed from the sum of these functions weighted by the spike rate occurring in each neuron [19]. In our TAB system, we have used a similar approach to decode the stimulus.

Accurate encoding of an input occurs when a population of neurons covers the entire range of the input variable. This is best achieved if the neuronal tuning curves are equally spaced, and may be imposed in a neural system by encoding the defined physiological properties of neurons in each population. However, the resulting costs to the system are unreasonably high. Instead, randomly chosen parameters from the distribution are likely to perform an equally good approximation [20]. Recently, Caron *et al.* showed the existence of such randomness in the olfactory system, where inputs from the glomeruli to individual Kenyon cells lack any organisation with respect to their odour tuning, anatomical features, or developmental origins [21]. In our TAB framework too, we have projected the input from the input layer neurons to the hidden layer neurons in a random manner. Random device mismatch cannot be avoided in smaller process technology, and instead we are exploiting it in the TAB framework to encode the input variable.

The TAB is a feed-forward network of three neuronal layers, namely input, hidden, and output, structured on the LSHDI principle [22] (Fig. 1). The input layer neurons are connected to a larger number of hidden layer neurons via fixed random weights. Consequently, the inputs are projected randomly and transformed to a higher dimensional feature space by the nonlinear hidden layer of neurons. In effect, the input data points, which are not linearly separable in their current space, find a linear hyperplane in the higher dimensional space that approximates a desired function as a regression solution, or represents a classification boundary for the input-output relationship. The output layer neurons derive a solution by computing only a linearly weighted sum of the hidden layer values. These linear weights are evaluated analytically by computing the product of the desired output values and the pseudoinverse of the hidden layer neurons output [23]. The TAB also employs systematic offset ($O_i$, Fig. 1) and preferred direction (PE), both of which help to diversify the tuning curves of the hidden layer neurons. Preferred direction (PE) implies that the activation function for one half of the hidden layer neurons may be *tanh*, and -*tanh* for the other half, and this may be assigned deterministically or randomly. Previously proposed networks based on the LSHDI principle include the Functional–Link Net computing (FLNN) by Pao *et al* in 1992 [24], the Extreme Learning Machine (ELM) by Huang *et al* in 2006 [25], and the Neural Engineering Framework (NEF) [26], which performs spike-based computation and is quite popular in the neuromorphic engineering community.

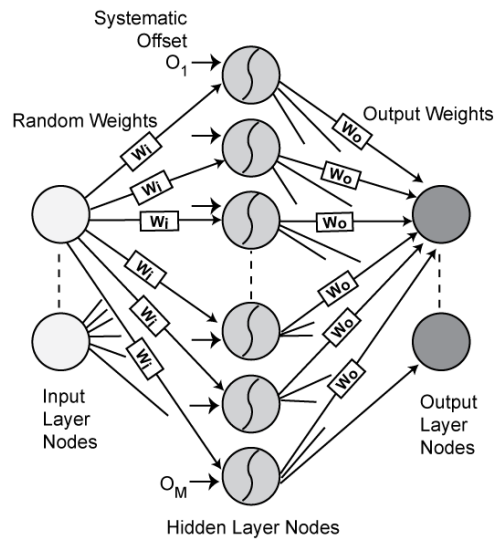

**Fig. 1. Architecture of the TAB framework.** The connections from the input layer neurons/nodes to the non-linear hidden neurons are via random weights and controllable offsets, $O_1$ to $O_M$. The hidden layer neurons are connected linearly to the outer layer neurons via trainable weights. The output neurons compute a linearly weighted sum of the hidden layer values. Adapted from [18]

## III. VLSI IMPLEMENTATION OF TAB

In order to demonstrate that the TAB is effective in smaller process nodes that are normally prohibitive to analogue design (at and beyond 65 nm) [27], we have designed the TAB prototype in a 65nm technology. Further, a substantial section of the TAB was designed using minimum feature sizes so as to maximise mismatch among transistor parameters. Differences among the hidden layer neuronal responses can be enlarged by using an additional distinct systematic offset for each hidden layer neuron. As a proof of concept we have implemented a single input-single output (SISO) version, with a single input

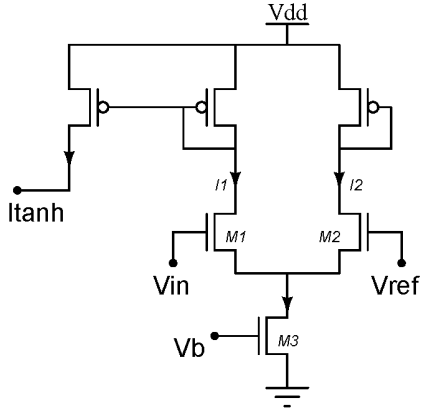

**Fig. 2. Hidden Neuron Block.** Schematic of the Hidden Neuron block that implements the *tanh* nonlinear activation function for the TAB framework. Adapted from [18]

voltage and a single output current. We elucidate below the VLSI implementation of the major building blocks of the TAB, namely the Hidden Neuron and the Output Weight.

### A. Hidden Neuron

Evidence has shown that neurons in a population respond to the same stimuli heterogeneously [28]. Each individual neuron encodes the input stimuli according to its own distinct tuning curve, which is found by presenting varied input stimuli to the neuron and recording its firing rate. We use a differential pair to implement a simple neuron model in the TAB. The differential pair performs a hyperbolic tangent (*tanh*) nonlinear operation on its input, similar to the sigmoidal tuning curve of the stereo V1 neurons in the cortex [29]. $M_1$ and $M_2$ constitute the differential pair in the circuit (Fig. 2). $V_{in}$ (input voltage) and $V_{ref}$ (constant reference voltage) are the gate voltages for $M_1$ and $M_2$, and influence the sharing of currents between them. If all the MOSFETs (metal-oxide semiconductor field-effect transistors) are operating in weak-inversion and at saturation, with the slope factor (*n*) ranging from 1.1 to 1.5, then the currents in transistors $M_1$ and $M_2$ can be approximated as:

$$I_1 = I_b[exp(V_{in}/nU_T)] / [exp(V_{in}/nU_T) + exp(V_{ref}/nU_T)] \quad (1)$$
$$I_2 = I_b[exp(V_{ref}/nU_T)] / [exp(V_{in}/nU_T) + exp(V_{ref}/nU_T)] \quad (2)$$

where, $I_b$ is the maximum bias current, $V_{in}$ is the ramp input voltage, $V_{ref}$ is the constant input voltage, and $U_T$ is the thermal voltage.

For ideal transistors, the output currents, $I_1$ and $I_2$, are a function of the input differential voltage between $V_{in}$ and $V_{ref}$ and their difference is identical to the *tanh* function. The current $I_1$ saturates to the maximum bias current if $V_{in}$ is higher than $V_{ref}$ by more than $4 \times U_T$ (~100 mV). The assumptions in (1) and (2) are true only for ideal transistors, and do not hold true for real circuits, resulting in deviation from the *tanh* curve. The current $I_1$ is copied to $I_{tanh}$ via a current mirror. The voltage at the $M_3$ transistor, $V_b$, sets the bias current (~few nanoamperes). In the TAB, each neuron has a distinct tuning curve depending on the process variations such as offset mismatch between the transistors in the differential pairs, bias current mismatch due to variability in $M_3$ and current mirror mismatch. Each neuron may receive a systematically different $V_{ref}$ in the TAB, which is a failsafe method to achieve a distinct tuning curve for each neuron. This may be required in the case of insufficient random variations, which is likely in higher feature size process technology.

### B. Output Weight

The output weight block connects the hidden layer and the output layer via linear weights. These are controlled by a 13-bit binary number that regulates the amount of current flowing from the hidden layer neurons to the output layer neurons. We have implemented binary weighted connections using a splitter circuit (Fig. 3) [30]. The output from the hidden neuron block, $I_{tanh}$, is the input current for the output weight block. $I_{tanh}$ is divided successively to form a geometrically-spaced series of smaller currents. A digital binary switch controls each current branch. A fixed fraction of the current is split off at each branch, and the remainder continues to the later branches. There are a total of N stages in the splitter circuit. The current at the $k^{th}$ stage is given by $(I_{tanh}/2^k)$. The master bias voltage $V_{gbias}$ is the reference voltage for the p-FET gates in the splitter [30]. Two transistor switches in the lower half of the R2R block act as a binary synapse for every branch, routing the branch current to either useful current, $I_{good}$, or to current that goes to ground, $I_{dump}$. $I_{good}$ is mirrored to generate a presynaptic current, $I_{out}$, for the output layer neuron. Internal shift registers of the hidden neurons are connected serially as a long chain. The shift registers are loaded with off-chip calculated weights and are used to regulate the current in the output weight block.

A TAB prototype was integrated on a 1 mm$^2$ die and fabricated in a 65nm process technology. This chip was designed for a single input and a single output configuration with 456 neuron blocks, the number of neuron blocks being constrained by the chip area. Each neuron block integrates a hidden neuron, an output weight block and a 13-bit shift register, which is used for loading the learnt weights. At a particular time, each neuron block receives the same input voltage, $V_{in}$, which is weighted by a random weight and a random offset arising due to process variations. Additionally, each neuron may exhibit a distinct reference voltage, $V_{ref}$, in the differential pairs of the *tanh* block. This leads to different differential voltages for each neuron block, and as a result different currents, $I_{tanh}$, are generated for each block. Each $V_{ref}$ is tapped from a long poly-silicon wire, the end points of which are connected to the top level voltage pins, $V_{ref1}$ and $V_{ref2}$. The poly-silicon wire behaves as a long distributed resistor element that acts as a voltage divider and generates different reference voltages, $V_{ref}$, for each neuron block. For each new input, the *tanh* block calculates $I_{tanh}$, which passes to the output weight block. In the output weight block, $I_{good}$ is mirrored to make $I_{out}$ (Fig. 3). $I_{out}$ is further routed to currents

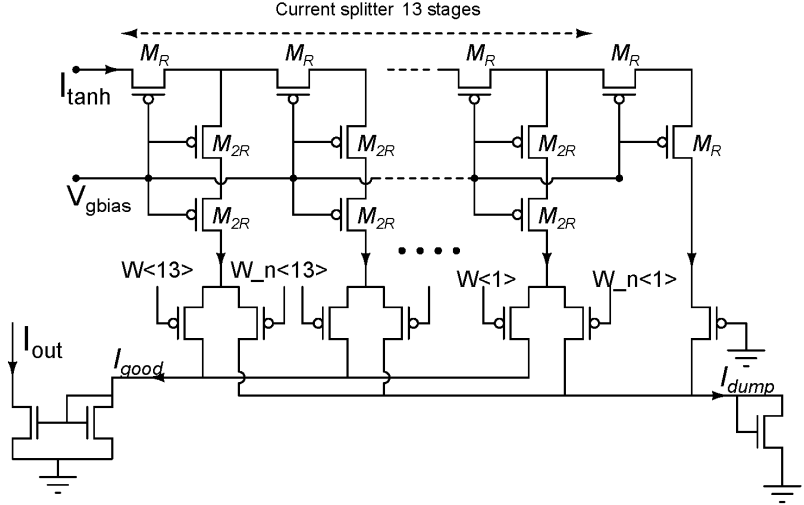

**Fig. 3. Output Weight Block.** Schematic of the Output Weight block, comprising a splitter circuit wherein $M_R$ and the two $M_{2R}$ transistors form an R2R network, which gets repeated 13 times in the block. The octave splitter is terminated with a single $M_R$ transistor. Adapted from [18]

$I_{outP}$ (positive current) or $I_{outN}$ (negative current), as determined by $signW$ signal, which indicates the polarity of the output weight connected between the hidden neuron and the output neuron. $I_{outP}$ and $I_{outN}$ currents of each neuron block are connected globally to each other, and they are summed up to provide the final current that is the final output of the TAB. We use an off-chip current-to-voltage converter and amplifier circuits to convert the final output current into voltage for ease of measurement.

## IV. LEARNING SET-UP

The algorithm used for the offline learning of the TAB IC is discussed here. In the TAB framework, learning is achieved by computing output weights to train the system for desired regression/classification tasks. The input layer of neurons connects to a much larger hidden layer using random weights. The output weights (between the large hidden layer and the linear output neurons) are estimated analytically by calculating the product of the pseudoinverse of the hidden layer activations with the target outputs [31].

Let us consider a three-layer feed-forward TAB network with $L$ number of hidden neurons. Let $G(.,.,.)$ be a real-valued function so that $G(w_i^{(1)}, b_i^{(1)}, o_i^{(1)}, x, d_i^{(1)})$ is the output of the $i$th hidden neuron, corresponding to the input vector $x \in \mathbb{R}^m$ and the random input weight vector $w_i^{(1)} = (w_{i1}^{(1)}, \dots w_{im}^{(1)})$, where $w_{iS}^{(1)}$ is the weight of the connection between the $i$th hidden neuron and $s$th neuron of the input layer, with random bias $b_i^{(1)} \in \mathbb{R}$, both arising due to random mismatch of the transistors. These random variables vary under a Gaussian distribution. Preferred direction (PE), denoted as $d_i^{(1)} \in [-1,1]$ is added to incorporate flexibility of changing the direction of the tuning curve either towards positive or negative values. PE assignment to the hidden neurons could be chosen either randomly or deterministically. Systematic offset $o_i^{(1)} \in \mathbb{R}$ is added to ensure that each neuron exhibits a distinct tuning curve, which is an essential requirement for learning in the LSHDI framework [18]. The output function $f(.)$ is given by:

$$f(x) = \sum_{i=1}^{L} w_i^{(2)} G(w_i^{(1)}, b_i^{(1)}, o_i^{(1)}, x, d_i^{(1)}) \qquad (3)$$

where, $w_i^{(2)} = (w_{1i}^{(2)}, \dots w_{ki}^{(2)}) \in \mathbb{R}^k$ is the weight vector where $w_{ji}^{(2)} \in \mathbb{R}$ is the weight connecting the $i$th hidden neuron with the $j$th neuron of the output layer. Here, $G(.,.,.)$ takes the following form:

$$G(w_i, b_i, x) = (g(w_i^{(1)}.x + b_i^{(1)} + o_i^{(1)})).d_i^{(1)} \qquad (4)$$

where, g: $\mathbb{R} \rightarrow \mathbb{R}$ is the activation function.

Suppose, for a training data set $\{(x_n, y_n)_{n=1,2..C}$, $x_n = (x_{n1}, \dots, x_{nm}) \in \mathbb{R}^m$ denotes the input vector, $y_n = (y_{n1}, \dots, y_{nk}) \in \mathbb{R}^k$ is its corresponding output vector, and $C$ is the total number of input data patterns. Let the values of the input weight vectors, $w_i^{(1)} \in \mathbb{R}^m$, the bias, $b_i^{(1)} \in \mathbb{R}$, be randomly assigned and $o_i^{(1)} \in \mathbb{R}$, be assigned systematically. Then, the standard TAB framework with $L$ number of hidden neurons approximates the input samples with zero error if and only if there exists $w_i^{(2)} \in \mathbb{R}^k$ such that:

$$y_n = \sum_{i=1}^{L} w_i^{(2)} G(w_i^{(1)}, b_i^{(1)}, o_i^{(1)}, d_i^{(1)}, x_n) \qquad (5)$$
$$\text{where, } n = 1,2,..C$$

The above set of equations can be rewritten in the following matrix form as:

$$HW^{(2)} = Y \qquad (6)$$

where,

$$H_{CxL} =$$
$$\left\{ \begin{matrix} G(w_1^{(1)}, b_1, o_1, d_1, x_1) \dots \dots G(w_L^{(1)}, b_L, o_L, d_L, x_1) \\ \vdots \qquad \qquad \qquad \vdots \\ \vdots \qquad \qquad \qquad \vdots \\ G(w_1^{(1)}, b_1, o_1, d_1, x_C) \dots \dots G(w_L^{(1)}, b_L, o_L, d_L, x_C) \end{matrix} \right\} \qquad (7)$$

$$W^{(2)}{}_{LxK} = \left\{ \begin{array}{c} w_1^{(2)} \\ \vdots \\ \vdots \\ w_L^{(2)} \end{array} \right\}, \qquad Y_{MxK} = \left\{ \begin{array}{c} y_1 \\ \vdots \\ \vdots \\ y_C \end{array} \right\} \qquad (8)$$

Here, the $i^{th}$ column of $H$ will be the output of the $i^{th}$ hidden neuron for all the input training data samples ($x_1,..., x_C$). Further, matrix $H$ need not be a square matrix. Under the assumption that the activation function $g(.)$ is infinitely differentiable, it has been shown that for fixed input weight vectors, $w_i^{(1)}$, and biases, $b_i^{(1)}$, $o_i^{(1)}$, the least squares solution $W^{(2)}$ for the matrix (6) is:

$$W^{(2)} = H^+ Y \qquad (9)$$

where, $H^+$ is the Moore-Penrose generalised pseudoinverse of the matrix $H$.

*Estimation of Output:* A network with the determined output weight vectors $w_i^{(2)} \in \mathbb{R}^k$ for the randomly chosen weight vectors $w_i^{(1)} \in \mathbb{R}^m$, random bias $b_i^{(1)} \in \mathbb{R}$ and systematic offset $o_i^{(1)} \in \mathbb{R}$ for $i = 1,2,...L$ will compute the estimated output value $\hat{y}$ for any input test sample $x \in \mathbb{R}^m$ using the following formula:

$$\hat{y} = \sum_{i=1}^{L} w_i^{(2)} \left( g(w_i^{(1)}.x + b_i^{(1)} + o_i^{(1)}) \right) . d_i^{(1)} \qquad (10)$$

## V. Constrained Algorithm

The pseudoinverse algorithm is a quick and easy method to estimate the parameters of a linear regression problem with a quadratic cost function. It is an unconstrained algorithm that can compute output weights in any range, and thus is not suited for hardware implementation of the TAB system. As shown in Fig. 3, the R2R block acts as a current divider, implying that the ratio of the output to the input current is always less than 1. Thus, we used the method of least squares to calculate the output weights, with additional constraints to calculate weights in the range of (-1, 1). As mentioned in the set up above, we collect matrix $H$ (in (7)) for all the training inputs. We created a function, *fcost_grad*, which calculates the squared error cost function *fcost* and gradient *fgrad* for a given training set. Our purpose is to estimate $W^{(2)}$ while minimising the cost function *fcost*. The function *fcost* is passed to the *fmincon* function of MATLAB as an argument along with the weight constraints. We define the cost function and gradient of the linear regression problem as:

$$fcost(W^{(2)}) = \\ (1/2C) \sum_{i=1}^{C} \sum_{i=1}^{L} (w_i^{(2)} g(w_i^{(1)}.x + b_i^{(1)} + o_i^{(1)}) . d_i^{(1)} - y)^2 \qquad (11)$$

$$fgrad(W^{(2)}) = \\ (1/C) \sum_{i=1}^{C} \sum_{i=1}^{L} (w_i^{(2)} g(w_i^{(1)}.x + b_i^{(1)} + o_i^{(1)}) . d_i^{(1)} - y) . x \qquad (12)$$

The function *fmincon* is an optimisation solver that finds the minimum of a constrained function. For linear regression, we want to minimise the cost function *fcost* with parameters $W^{(2)}$ using a given fixed training set.

## VI. Results

### A. Analysis of hidden nodes and output weights

The size of a circuit grows linearly with the number of hidden nodes and the number of bits in the output weights. We first examined the number of hidden nodes and bits needed in our system using software simulations. The optimum number of hidden nodes and bits was found to vary with the desired target function, with some functions, such as $sinc(x)$, proving much more difficult to learn than others, such as $x^2$. Fig. 4A shows the RMS error between the target function and the learned function as a function of the number of hidden neurons, for both $y = sinc(6\pi x)$, and $y = x^2$. The mean RMS error was calculated for 10 simulations with different random weights from the input to the hidden nodes. In these simulations, we used 13-bits for the output weights. It can be seen that the standard deviation is quite high for low number of hidden nodes, but for a larger number of hidden nodes the result becomes largely independent of the random weights. It is also clear that the $y = sinc(6\pi x)$ function needs significantly more hidden nodes than $y = x^2$ to be learned accurately.

The RMS error as a function of the number of bits used for the output weights, for $y = sinc(6\pi x)$ for a network with 100 hidden nodes is shown in Fig. 4B. Again, these are the results of 10 simulations with different random weights from the input to the hidden nodes. Clearly, the higher the resolution in the output weights, the closer the digital weights can approach the weights found by the offline learning (which are real-valued numbers) and the better the function can be implemented. From about 8-bits onwards, the standard deviation is negligibly small, and the RMS error becomes almost totally independent of the random weights. Increasing the number of bits per weight is a matter of diminishing returns, and 11-bits seem sufficient, even to learn this difficult function.

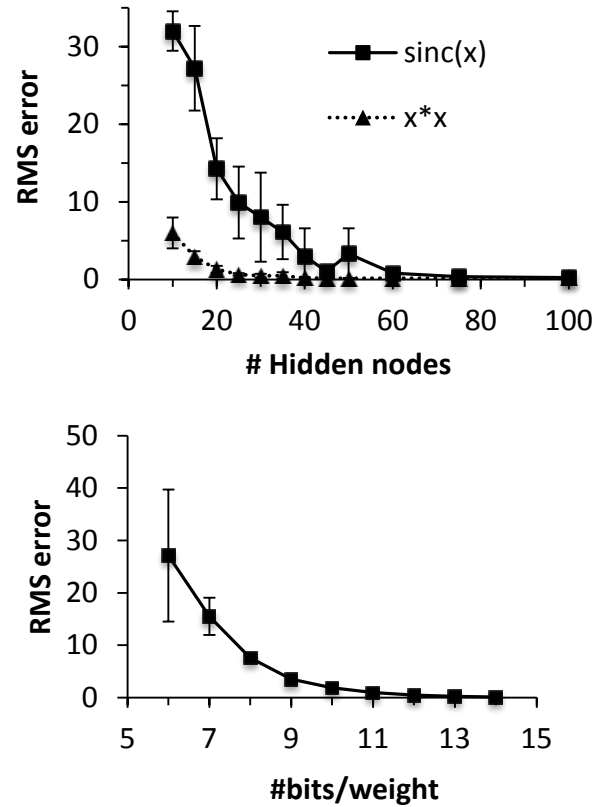

**Fig. 4. A.** Plot of the RMS error between the target function and the learned function versus the number of hidden nodes. The error bars show the standard deviation. **B.** The RMS error versus the number of bits/weight used for the output weights for the function $y = sinc(6\pi x)$. The error bars show the variance.

### B. TAB learning in software simulation

Here, we show the software simulation results of the TAB network using a single input and a single output configuration. We built the TAB network with 50 hidden neurons with 13bits output weight, and tested its ability to learn different functions such as *sin* and *square*, using constrained offline learning. In the TAB, we used the *tanh* function as the nonlinear activation function (tuning curve) for each hidden neuron. We used the offline learning setup as mentioned in section IV, and the *fmincon* solver from MATLAB to calculate the output weights externally. We presented the training data to the network, each training pair containing an input, *x*, and an output, *y*. Each input training value is randomly and systematically offset and multiplied by random weights of a Gaussian distribution for each hidden neuron, and is projected randomly to 50 hidden neurons in this manner. For every input data point, we collected the response of the hidden neurons and created a matrix *H* as shown in (7). We used the constrained algorithm to calculate the output weights. In the testing phase, we presented the test input to the network and obtained a predicted output. We show that the TAB system is able to learn the various functions successfully (Fig. 5). We normalised our input range between -1 to +1 V to simplify the function argument. From the simulation results, we can see that the learned function closely matches with the target function. Fig. 5A and 5B show the results from training the network for the *sin* and *square* functions, respectively.

### C. Neuron characterisation in the TAB IC

A TAB prototype was fabricated in the 65nm process technology with 456 neuron blocks. Here, we characterise the tuning curve of each neuron to analyse the mismatch and differences between the tuning curves of the hidden neurons without any systematic offset by connecting $V_{ref}$ node (Fig.2) of each hidden neuron to the same voltage,. Learning is better if there is a high diversity between the tuning curves of neurons. As shown in Fig. 6, we obtained heterogeneity in the neuronal tuning curves due to random device mismatches and process variations in the fabrication. Each neuron block contains a hidden neuron and output weight block (section III). Also, each neuron block contains a 13-bit shift register. The shift registers of all the hidden neurons are connected serially as a long chain. Due to the large number of hidden neurons, it is not feasible to have dedicated output ports to probe the output current for each hidden neuron. Therefore, the output current of each hidden neuron is probed indirectly through the 'OUT' port of the IC one-by-one. The output weight block behaves as a current splitter, i.e., if all the bits of a weight are one, the output current of the output weight block would be almost the same as its input current, and if all the bits of the weight are zeros, the output weight block will produce nearly zero output. We characterised each neuron sequentially. We loaded the output weight between the hidden neuron of interest and the 'OUT' port to be 13'h1FFF, and all other weights between the remaining hidden neurons and OUT port to be 13'h0000. The MSB (most significant bit, 13[th] bit here) represents the sign of the output weight, where '1' represents negative weight. Then, we provided the ramp input to the TAB and measured the current at the output port. The input to the TAB and the outputs of the tuning curves of a few neurons were plotted in Fig. 6A and Fig. 6B respectively. We provided 0.25V to the bias transistor M3 in the hidden neuron circuit (Fig. 2), and we expected ~8-10 nA output current according to circuit simulations. We collected tuning curves of all the neurons and plotted the statistical variation in amplitude and offset of the tuning curves (Fig. 6C, D).

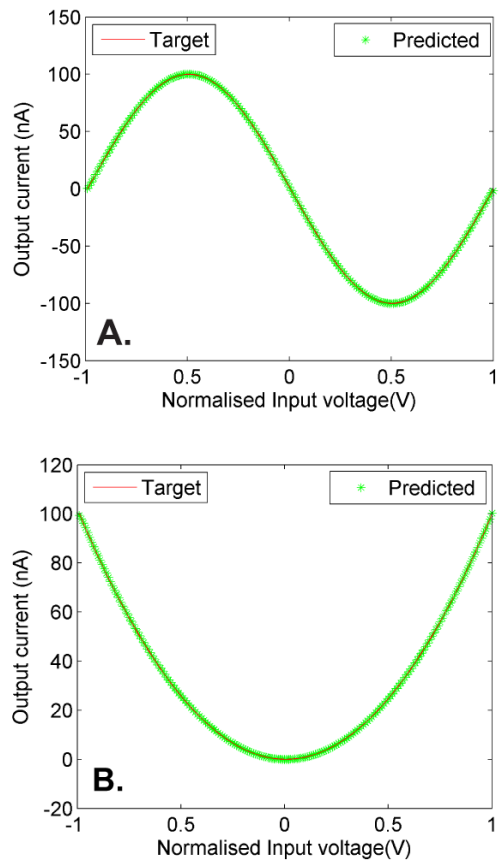

**Fig. 5.** Learning curves for the regression functions **A.** *sin*, **B.** *square*. The *red* curve represents the target function, and the *green* curve represents the learnt function.

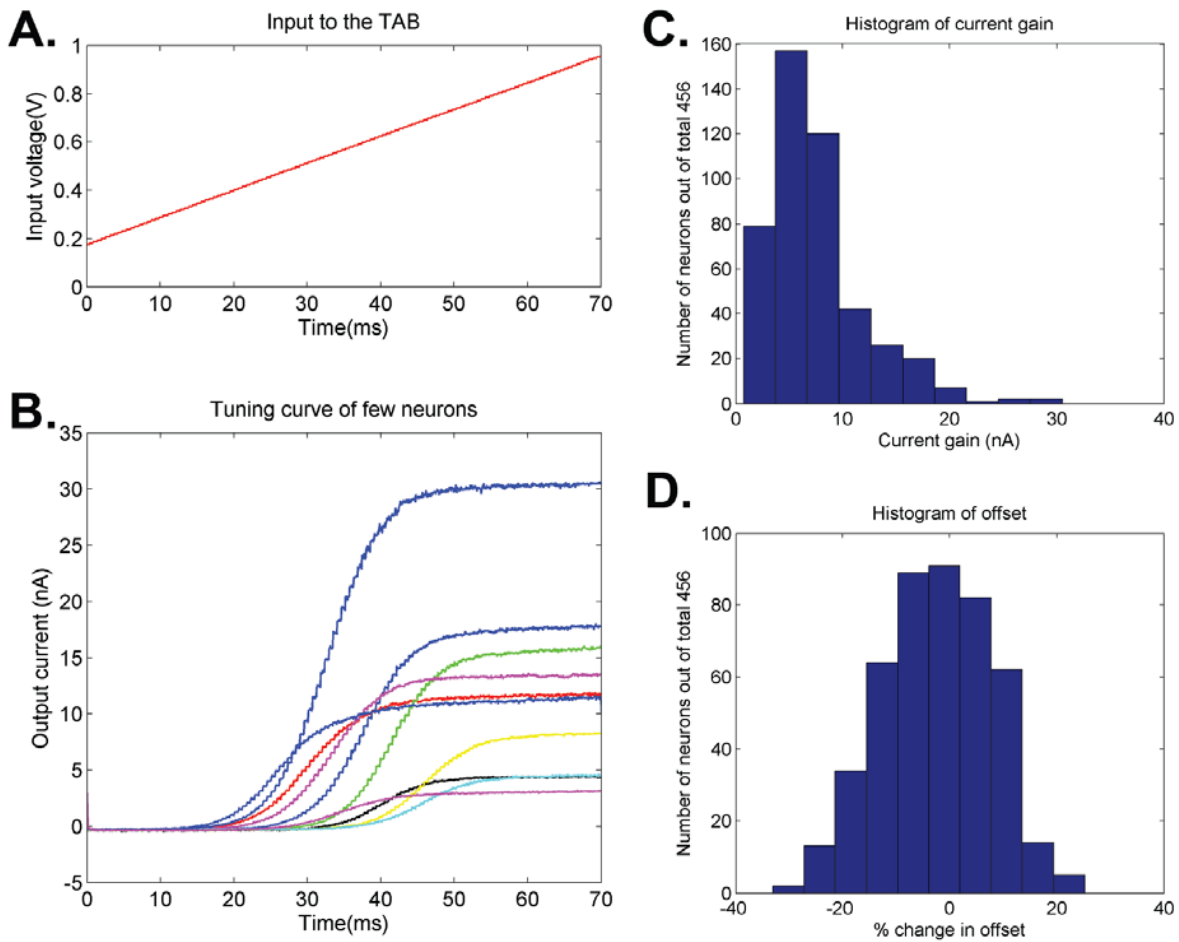

**Fig. 6.** Hidden neurons characterisation results for a total of 456 neurons. **A.** Input to the system (*red* line). **B.** *tanh* tuning curves of a few neurons to show the variation in offset and the minimum current value. **C.** Variation in the current amplitude of tuning curves of all the 456 neurons. **D.** Random offset present in the chip across all the 456 neurons.

### D. Learning capability of the TAB IC

In this section, we show the learning capability of the TAB IC. This chip was designed for a single input and a single output. Each neuron block contains a hidden neuron block and an output weight block, as explained in section III. Also, each neuron block contains a 13-bit shift register that is used to load the offline trained weights. To calculate the output weights externally, the output of hidden neurons for a given input sample needs to be probed. For $L$ hidden neurons and $C$ distinct input training data $(x_1..x_C)$, the matrix $H_{CxL}$ as shown in (7) is obtained, and the value of weights is calculated using the offline learning setup as described in section IV. We collected the tuning curves of all the hidden neurons sequentially by measuring the current at the output port, as described above in section VI.D. After collecting all the data, the output weights were calculated using the constrained algorithm. We trained the TAB IC for various regression tasks such as *sin, cube, square* functions as shown in Fig. 7. The output for each of the functions matched well with the desired output, indicating that the TAB IC learnt the specific function.

### E. Encoding capacity of the TAB IC due to random mismatch

We have seen the learning capability of the TAB IC consisting of 456 hidden neurons. Variations in the neuronal tuning curves have been shown in Fig. 6, which are the result of random device mismatch. These variations make the tuning curves different from each other, which is key in learning any regression or classification task. In this section, we estimate the actual number of hidden neurons needed to learn a regression function, for example the *sinc* function. Importantly, we will measure variations in the RMS error from using different combinations of tuning curves encoding the input. We use a few tuning curves out of the 456 hidden neurons from the TAB chip in response to the ramp input for learning the *sinc* function. Fig. 8 shows the distribution of the errors for 100 different combinations of 40 hidden neurons chosen randomly out of a total 456 neurons. The results suggest that some combinations of hidden neurons encode better than others. Moreover, combinations of 40 hidden neurons out of 456 are sufficient to learn this function with a average RMS error of 4.9%, and a standard variation of 0.95% with respect to the RMS of the target signal. It can be concluded that the TAB chip has a large encoding capacity as a result of the diversity of neuronal tuning curves, which in turn arises from random device mismatches. The TAB is thus a unique framework that overcomes the limitations of random device mismatches and employs them to its advantage.

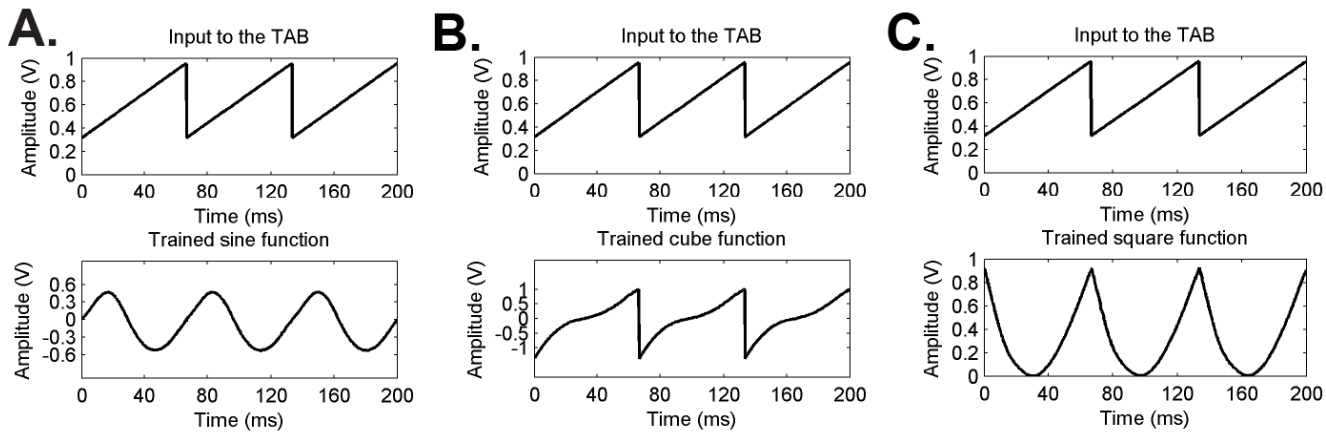

**Fig. 7.** Learning in the TAB chip for the functions: **A.** *sine*, **B.** *cube*, **C.** *square*. Top and bottom graph in each figure represent input to the TAB and trained output from the TAB.

## VII. CONCLUSIONS

In this paper, we have presented a novel neuromorphic TAB architecture inspired from the phenomenon of population coding present in the nervous system, and exploits random device mismatch (fixed-pattern mismatch) and variability in the fabrication process to perform reliable computation. We have presented measurement results of our first prototype IC designed in 65nm for a single input and a single output configuration of the TAB system. To our knowledge, we have developed the first IC that exploits random device mismatch to its advantage. The TAB also incorporates systematic offset as a failsafe method to spread the tuning curves of the neurons. Systematic offset may be required when there is insufficient random variation among transistors to produce a distinct tuning curve for each neuron, which is highly likely in higher feature size process technology. We have also shown the learning capability of the TAB system for various regression tasks.

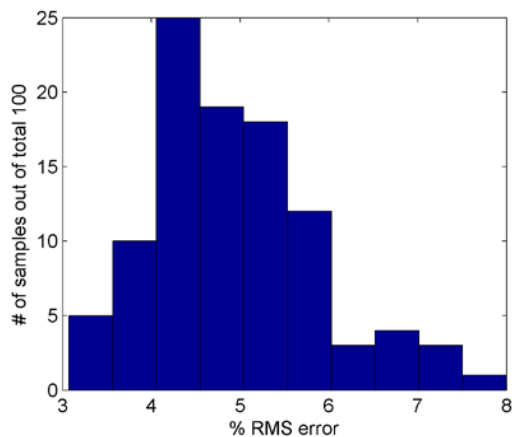

**Fig. 8.** Distribution of error while training the TAB for learning the *sinc* function using 40 hidden neurons randomly sampled 100 times in total from a pool of 456 neurons.

The TAB is inspired by neural population coding which is very robust in the face of damage of a few neurons, and does not have a disastrous effect on the encoded representation as the information is encoded across many neurons. The TAB system is designed using neuromorphic principles based on stochastic computation, which has the advantages of low power consumption since the circuit is operating in the subthreshold regime, fault tolerance, adaptability to local change and the ability to learn. We envisage the TAB to overcome the limitations of analogue IC design at low process nodes and drive the integration process with digital blocks in the same circuit and process node. This may find applications in analogue/digital converters (ADCs) and digital-to-analogue converters (DACs) for submicron mixed signal chips such as those used in mobile processor chips and data acquisition chips.

The main significance of our TAB is that it solves the problem that increased device mismatch in modern IC manufacturing technologies causes for analogue design. It works better with a considerable amount of device mismatch, and accurate mappings from input values to output values can be obtained without needing to engineer the effect of device mismatch out of the circuit, as is done currently. A further significant advantage of this approach is that the same TAB may be reused for many different purposes once manufactured, and the same architecture may be used in different manufacturing technologies. This will lead to a significantly reduced design cycle for analogue circuits, with an associated reduction in design cost, and a speed-up of technological progress. The TAB may also be (re-)trained 'on the job'. This could be a major advantage in systems where the input-output mapping of a TAB needs to be changed because of changes in the system. An example would be in a communication system where a TAB is used as a filter to process the analogue signal before digitisation, in which the communication channel changes over time. The TAB can be re-trained with the communication channel in the loop to compensate for the changes. Furthermore, as the TAB framework desires large random mismatch among devices, and as mismatch is inversely proportional to device area, it could lead to significant reductions in chip area and manufacturing costs. Also, the failure of a few neurons would not affect the performance of the TAB as information is encoded into a large ensemble of neurons. This property would help significantly in improving the overall yield of the chip production. Future work will aim to design TAB chips for multiple inputs.